\documentclass[conference]{IEEEtran}
\IEEEoverridecommandlockouts
\usepackage{cite}
\usepackage{amsmath,amssymb,amsfonts}
\usepackage{algorithmic}
\usepackage{graphicx}
\usepackage{float}
\usepackage{caption}
\usepackage{placeins}
\usepackage[table]{xcolor}
\usepackage{textcomp}
\usepackage{subcaption}
\usepackage{mwe}
\usepackage{url}
\usepackage{multirow}
\usepackage{booktabs}
\usepackage{xcolor}
\usepackage{textcomp}
\usepackage{stfloats}
\usepackage{url}

\def\BibTeX{{\rm B\kern-.05em{\sc i\kern-.025em b}\kern-.08em
    T\kern-.1667em\lower.7ex\hbox{E}\kern-.125emX}}

\begin{document}

\title{Spinal ligaments detection on vertebrae meshes using registration and 3D edge detection}

\author{ Ivanna Kramer$^{1}$, Lara Blomenkamp$^{1}$, Kevin Weirauch$^{2}$, Sabine Bauer$^{3}$ and  Dietrich Paulus$^{1}$
\thanks{*This work was not supported by any organization.}
\thanks{*This paper is accepted at 46th Annual International Conference of the IEEE Engineering in Medicine and Biology Society, 2024.}
\thanks{$^{1}$All authors are with the Active Vision Group, Institute for Computational Visualistics, University of Koblenz, Germany.}
\thanks{$^{2}$ This author is with the Intelligent Autonomous Systems Group, Institute for Computational Visualistics, University of Koblenz, Germany}%
\thanks{{$^{3}$ This author is with  Institute for Medical Technology and Information Processing, University of Koblenz, Germany.}%
\newline
{Corresponding email - \tt\small ikramer@uni-koblenz.de}}
    }


\maketitle

\begin{abstract}

Spinal ligaments are crucial elements in the complex biomechanical simulation models as they transfer forces on the bony structure, guide and limit movements and stabilize the spine. The spinal ligaments encompass seven major groups being responsible for maintaining functional interrelationships among the other spinal components. Determination of the ligament origin and insertion points on the 3D vertebrae models is an essential step in building accurate and complex spine biomechanical models. In our paper, we propose a pipeline that is able to detect 66 spinal ligament attachment points by using a step-wise approach. Our method incorporates a fast vertebra registration that strategically extracts only 15 3D points to compute the transformation, and edge detection for a precise projection of the registered ligaments onto any given patient-specific vertebra model. 
Our method shows high accuracy, particularly in identifying landmarks on the anterior part of the vertebra with an average distance of 2.24 mm for anterior longitudinal ligament and 1.26 mm for posterior longitudinal ligament landmarks. The landmark detection requires approximately 3.0 seconds per vertebra, providing a substantial improvement over existing methods. Clinical relevance: using the proposed method, the required landmarks that represent origin and insertion points for forces in the biomechanical spine models can be localized automatically in an accurate and time-efficient manner. 
\end{abstract}



\section{Introduction}

\begin{figure}[htb]

 \centering
  \centerline{\includegraphics[height=4.5cm]{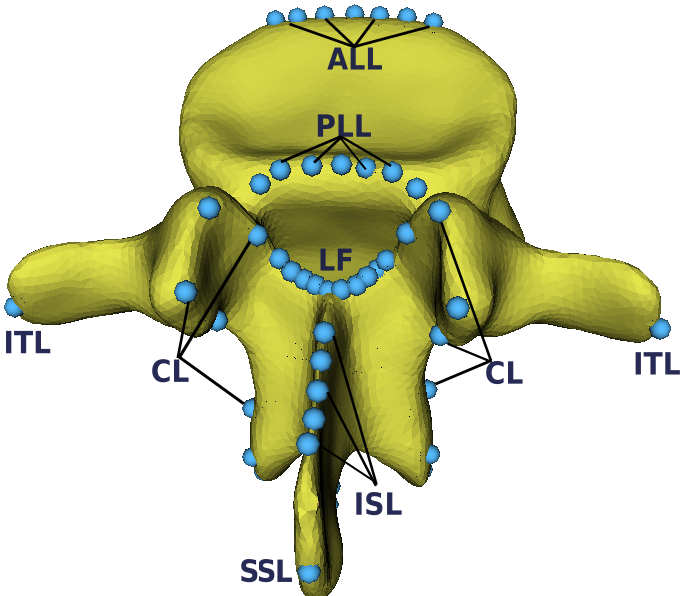}}
\caption{Anatomically grouped 3D ligament landmarks annotated on the artificial L3 vertebra model \cite{panjabi1980basic}. The ligaments PLL, ALL, ITL, FL, and CL, which run over a particular surface area, are divided into different ligament bundles.}  \label{fig:sample_ligs}

\end{figure}

Spinal ligaments are bands of fibrous connective tissues which connect individual bones and, together with the spinal muscles, provide stability to the vertebral column during such excessive movements as hyper-extension or hyper-flexion and contribute in resistance of the spine to tensile loads \cite{panjabi1980basic}. There are seven major ligament groups that are responsible for maintaining functional interrelationships among the other spinal components: anterior longitudinal ligament (ALL), posterior longitudinal ligament (PLL), capsular ligament (CL), ligamentum flavum (LF), and interspinous and supraspinous ligaments (ISL and SSL), and intermuscular transverse ligament (ITL) \cite{panjabi1980basic}. The origin and insertion points of the corresponding ligament groups can be seen Fig. \ref{fig:sample_ligs}. 

 The clinical relevance of automatic ligament detection arises directly from the need for improved biomechanical simulations for which it is mandatory to quantify the loads on the spinal structures. Better biomechanical simulations can then lead to greatly improved spine treatment strategies due to the ability of observing long-term effects without health risks for patients.
 
A biomechanical simulation model of the spine consists of various components, such as the vertebral bodies, facet joints, muscles and intervertebral discs, as well as the ligament structures. In order to adequately represent the specific biomechanical properties of the ligaments in a simulation model, the exact location of the attachment points of the ligaments is necessary, as the ligament forces are applied and transmitted via these attachment points.
Therefore accurate modelling of the ligaments in the spinal biomechanical models is crucial for understanding the force distribution and motion sequences within the spine \cite{damm2020lumbar}, \cite{juchem2009mbs}. If the sufficiently precise localization of the ligament origin and insertion points is not guaranteed, non physiological lever arms can lead to torques that negatively affect the quality of the simulation results. 

Biomechanical models of the spine that represent reality with sufficient accuracy are increasingly complex, incorporating numerous ligaments with specific origins and insertions \cite{Kramer21}. The process of manually identifying and marking ligament landmarks on 3D vertebrae models is very time-consuming. Furthermore, the manual annotation of ligament attachment points is prone to human error and can vary significantly between different annotators. To reduce this error potential and the time required to create biomechanichal models, we seek to automate this process. Additionally this would allow for easier scaling of the model. 


In this paper, we propose a novel approach to detect ligaments origin and insertion points on 3D models of vertebra. Detecting the ligaments on the 3D vertebra meshs does not require a medical image and works on 3D spinal models obtained from different sources, i.e. from medical images as well as from optical systems that can reconstruct the human spine in motion \cite{degenhardt2020reliability}.
Our contribution in this paper resides in the following aspects: (1) we propose a fully automated pipeline to detect a large number (66) of ligament attachment points on the spinal meshes. (2) We enhance the reproducibility of our proposed method by releasing two datasets of 3D vertebra meshes, complete with manually annotated ligament points by experts from the field. The initial dataset includes 17 thoracic and lumbar artificial vertebrae models. Our second dataset is built from the publicly available VerSe 2021 \cite{sekuboyina2021verse} dataset, where we have randomly chosen the 3D vertebra models and provided them with ground truth values for ligaments. 
Finally, we also publish a ready-to-use 3D Slicer\footnote{\url{https://github.com/VisSim-UniKO/Spinal-Ligament-Detection}} plugin that implements our method.

\begin{figure*}[ht!]
    \centering
    \includegraphics[height=0.42\textwidth]{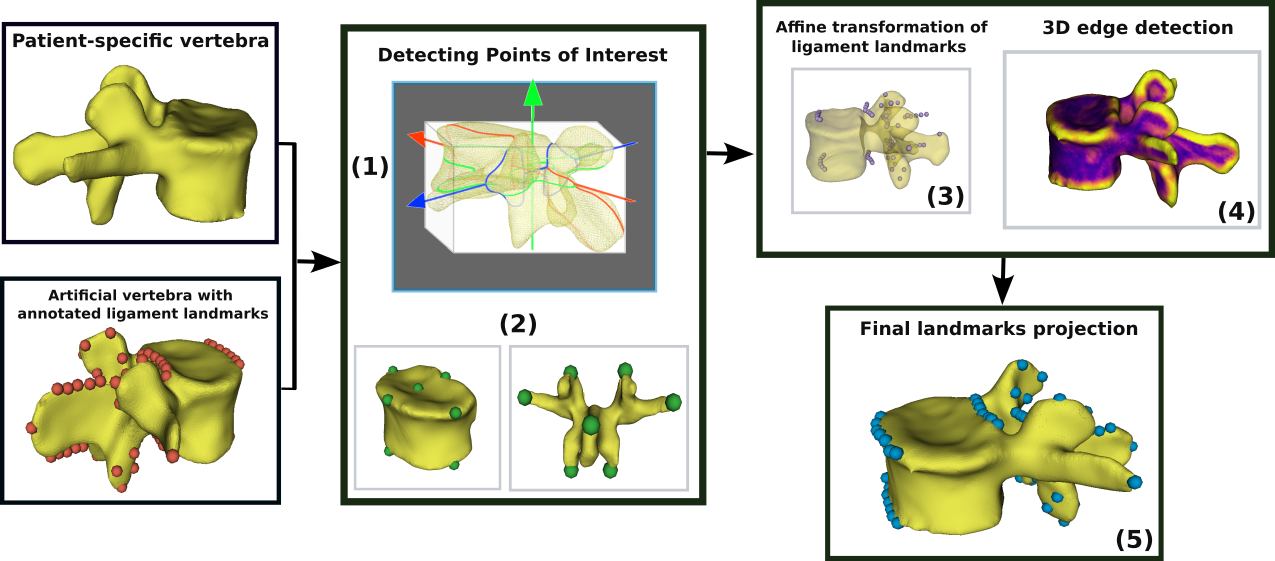}
    \caption{Workflow of the proposed method: Starting with annotated ligament landmarks on an artificial vertebra and the model of a patient-specific vertebra (left) we first detect 15 Points of Interest (PoIs) as the anatomical extrema/most outer points (middle). Using these points we register the annotated ligament landmarks to the patient specific vertebra (3). Since the landmarks might still be off due to the patient specific vertebra geometry, we now project them to the surface using edge detection (4). The result is (5), in which all landmarks are aligned with the patient specific vertebra geometry.}
    \label{fig:pipeline}
\end{figure*}

\section{Related Work}
In recent years, several studies addressed automatic determination of knee ligaments, however only a few publications focused on the detection of the attachment points for the spinal ligaments: automatic segmentation of knee ligaments was proposed using 
various methods like the Active Contour Method \cite{vinay2014active},
Deep Learning \cite{jonmohamadi2020automatic} or a  modified fuzzy C-means algorithm
\cite{zarychta2015feature}, though these methods do not consider 
the detection of origins and insertions for the segmented ligaments.  Ascani et al. \cite{ascani2015procedure} proposed a method for automatic
search for the origin and insertion points of the four knee ligaments
from computed tomography (CT) images using atlas-based registration.
The procedure was validated on 14 volumes containing 7 CT and 7 MRI datasets 
with pathological knees, where the ground truth landmarks were set by 
four different operators. The average difference between ground truth
data and validation was reported to be  $2.1\pm 1.2$ mm and 
$2.7\pm1.0$ mm for different ligaments.  Recently AL-Dhamary et al. \cite{ibraheem2019automatic}  proposed a semi-automatic technique for identifying the origin points of ligaments in the cervical spine. This method utilizes atlas-based registration of medical images, similar to the approach in the preceding study. Initially, the user identifies the relevant vertebra, which is followed by the segmentation of the vertebrae. Subsequently, the cervical ligament attachment points are determined. Lerchl et al., in their work in \cite{lerchl2022validation}, introduced an automated process for spinal segmentation that involves detecting ligament landmarks on 3D models of vertebrae as a final step. Their methodology involved subdividing the vertebrae into four sub-models and identifying the ligament attachment points as the most lateral and sagittal and minimum vertices within these sub-models.

Our approach presented in this paper is distinguished from the existing studies by applying an automatic method to 3D models of vertebrae rather than directly on medical images. This allows for the determination of ligament landmarks on vertebrae meshes generated from various systems, without being limited to the outputs of medical imaging scanners. Moreover, our technique focuses on identifying ligament landmarks based on their anatomical positions on the vertebrae, rather than merely finding the extreme points (maximum or minimum) within sub-sections of the vertebrae. This results in a more natural representation of spinal ligaments, enhancing the quality of the spinal models.


\section{Materials and Methods}

The goal of the proposed pipeline is to automatically find three-dimensional ligaments origin and insertion points, referred throughout this paper as ligament landmarks, on the patient-specific vertebra models. We use previously manually annotated artificial vertebrae models and register them to the patient-specific geometries in a time-efficient manner. The proposed pipeline is depicted in Fig. \ref{fig:pipeline}. The workflow of our method consists of several steps. In the first step (Fig. \ref{fig:pipeline},~Step (1)) we determine the local coordinate systems of both input meshes, which enables us to define anatomical planes for the corresponding vertebrae. Using these planes, we divide the vertebrae geometry into symmetrical parts to detect 15 points of interest (PoIs) located at the outermost geometric extrema  as depicted in Step (2) of Fig. \ref{fig:pipeline}.

An affine transformation is calculated from the two sets of detected PoIs. Using this transformation, we align the ligament landmarks with the patient-specific vertebral model. Because the patient-specific vertebral models differ from the ideal artificial vertebrae, the initial transformation of the ligament landmarks does not fit perfectly, which is shown in Fig. \ref{fig:pipeline}, ~Step (3). To accurately adjust the positions of the ligaments and map them onto the edges of the vertebral surface, reflecting their inherent anatomical locations, we perform 3D edge detection on vertebral models. Once, the edges are found (\ref{fig:pipeline}, ~Step (4)), the transformed landmarks are projected to their final positions using specific pre-defined rules for each group of ligaments. In the following we will explain this pipeline in further details.

\begin{figure*}[t!]
    \centering
    \includegraphics[width=1\textwidth]{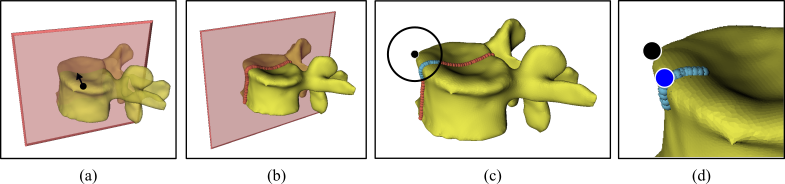}
    \caption{Workflow of the ligament landmark projection: (a) We calculate the center of mass of the initial landmarks and determine a plane oriented as in the picture intersecting the center of mass, (b) the intersection between the vertebra and the plane is detected, (c) within a fixed radius around each ligament landmark we determine so-called edge values in a similar manner to \cite{ahmed2018edge},  (d) the intersection point with the highest edge value is the most promising match for the ligament landmark.}
    \label{fig:projection}
\end{figure*}

\textbf{Detecting Points of Interest}: In this step of the method we detect a set of specific points of interest  on both the patient-specific and the artificial vertebrae. To identify the PoIs, we initially compute the object coordinate system for each vertebra, following the method described in \cite{kramer2023VertebraeMeasurements}. Subsequently, we establish two cutting planes based on the computed coordinate axes: the frontal plane, which divides the vertebrae into anterior and posterior sections, and the sagittal plane, which separates the mesh into right and left portions. These intersections precisely coincide with the anatomical extremities, or, from a geometrical perspective, the outermost points of the vertebrae meshes. For each vertebra, we identify exactly 15 such points, as illustrated in Step (2) of Fig. \ref{fig:pipeline}, where they are highlighted in green. We use the PoIs to calculate the affine transformation of the corresponding pair of vertebrae.

\textbf{Affine transformation}: Let $P_a, P_p$ denote the set of PoIs from the artificial vertebra model and the patient specific vertebra respectively. Assume that $\vert P_a\vert = n = \vert P_p\vert$. We seek an affine transformation $f: P_a \rightarrow P_p$ that minimizes $\sum_{i=1}^n \Vert P_a^i - P_p^i \Vert^2$ where $P_x^i$ denotes the i-th PoI within either point sets. This is a classical problem of finding the least squares minimizer.
Using the method of Horn we are able to find this transformation $f$ by translating the problem from the $3$D-coordinate system to a problem about quaternions and solving a Eigenvector problem for a $4\times 4$ matrix \cite{horn1987closed}.

\textbf{Finding 3D Edges:} For each point of the patient-specific mesh, we estimate an edge value with a method similar to \cite{ahmed2018edge}.
Using a KD-Tree, we determine points in the local neighborhood of on the mesh  in a defined radius  and calculate their centroid.
The distance from the point to this centroid evaluates the symmetry in the local neighborhood and is used to assign high probability values to the points that are located on an edge (see Fig.~\ref{fig:pipeline},~Step~4).


\textbf{Projecting Ligaments:} The projection of the ALL and PLL ligaments is shown in Fig. \ref{fig:projection}. For the other ligament groups this approach can be applied analogously. Essentially, for each ligament landmark we select the intersection point with the highest edge value to get the best projection. This way the landmarks fit specifically to the patients spine model.

\begin{table*}[h]
\centering
\caption{Comparison of different landmark registration methods. The error values are measured as average distances [mm]  of the detected ligament landmarks to the groundtruth.}
\begin{tabular}{lccccccccccc}
\toprule
\textbf{Method} & \textbf{ALL[mm]} & \textbf{PLL[mm]} & \textbf{ISL[mm]}  &  \textbf{CL[mm]} & \textbf{ITL[mm]} & \textbf{SSL[mm]} & \textbf{LF[mm]} & \textbf{Average[mm]} & \textbf{RMSE[mm]}   & \textbf{Time [s]} \\
\midrule
\hspace*{-0.2cm}
\textbf{Our 15 PoIs}         
    &    \textbf{2.24}
    &    1.68
    &    5.06
    &    4.88
    &    \textbf{2.39}
    &    \textbf{5.73}
    &    3.51
    &    3.64
    &    3.98
    &    3.21\\
    \hspace*{-0.2cm}
\textbf{Our 8 PoIs }
    &    \textbf{2.24}
    &    \textbf{1.26}
    &    7.33
    &    4.59
    &    32.07
    &    84.90
    &    4.07
    &    19.49
    &    6.14
    &    \textbf{3.0}\\
\hspace*{-0.2cm}
\textbf{ALPACA \cite{Porto2021ALP}}
    &    2.59
    &    1.96
    &    \textbf{3.96}
    &    \textbf{2.46}
    &    3.63
    &    6.67
    &    \textbf{2.27}
    &    \textbf{3.36}
    &    \textbf{3.53}
    &    77.6
    \\
\bottomrule
\vspace{0.1cm}
\end{tabular}

\label{tab:results_evaluation}
\end{table*}






\section{Experiments and Results}

\begin{figure}[htb]
    \centering
        \begin{subfigure}{0.24\textwidth}  
        \centering 
        \includegraphics[width=\textwidth]{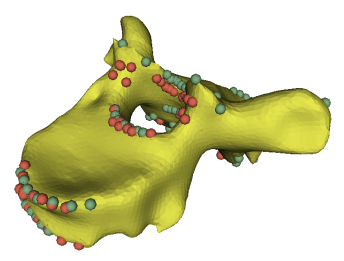}
        \caption{}
    \end{subfigure}
    \begin{subfigure}{0.24\textwidth}   
        \centering 
        \includegraphics[width=\textwidth]{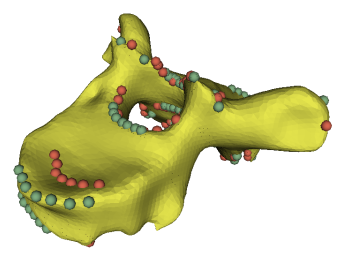}
        \caption{}
    \end{subfigure}
\caption{The ligament landmarks (red) are detected on the fractured vertebra model by (a) our method and (b) ALPACA \cite{Porto2021ALP} in comparison to the ground truth (green).}  
\label{fig:sample_res}
\end{figure}

As a performance metric for our method we used the distance of the average landmark from the ground truth. For this we calculated the average landmark for both our ground truth as well as for the landmarks produced by our method. We then determined the euclidean distance between these two points.
Additionally, we calculated the Root Squared Mean Error (RMSE)  to assess the precision between the detected and the ground truth landmarks in a 3D space. Average computational time (in seconds) per vertebra was reported as well.
In the conducted experiments we used two variations of our method, one with the 15 PoIs ("Our 15 PoIs") and one with 8 PoIs ("Our 8 PoIs"), that were detected on the upper and lower endplates of the corresponding vertebral bodies. We selected ALPACA \cite{Porto2021ALP} to be our baseline method.

\begin{figure}[htb]
    \centering
        \begin{subfigure}{0.24\textwidth}  
        \centering 
        \includegraphics[width=\textwidth]{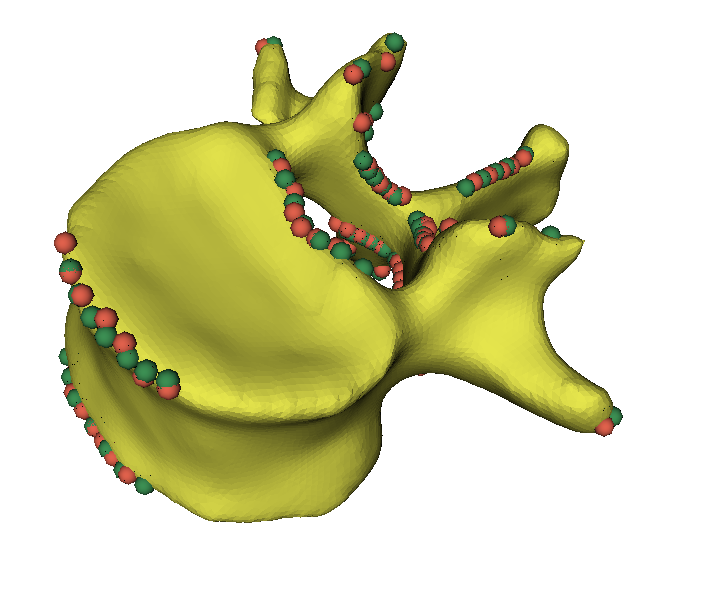}
        \caption{}
    \end{subfigure}
    \begin{subfigure}{0.24\textwidth}   
        \centering 
        \includegraphics[width=\textwidth]{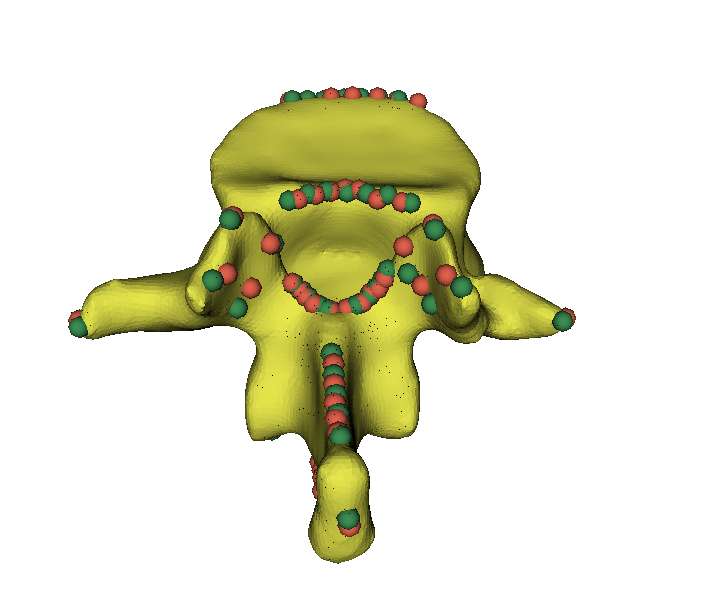}
        \caption{}
    \end{subfigure}
\caption{The ligament landmarks in the anterior (a) and posterior (b) parts of the healthy L1-vertebra are detected by our method (red) and compared to the ground truth (green).}  
\label{fig:sample_res_healthy}
\end{figure}

To evaluate the proposed pipeline, we used the segmented vertebrae from a public VerSe 2021 dataset \cite{sekuboyina2021verse}, and manually annotated 30 of them from thoracic and lumbar spines. 

Table \ref{tab:results_evaluation} presents a comparison of experimental results for different detection methods. It can be seen from the results, that our method (with 15 PoIs) shows slightly lower accuracy across the most posterior ligaments (all except ALL, PLL, ITL and SSL) with the highest error observed in the supraspinous ligament (SSL) at 5.73 mm and the lowest in the posterior longitudinal ligament (PLL) at 1.68 mm.
The overall average error is 3.64 mm with an RMSE of 3.98 mm, indicating consistent error distribution across the landmarks.

The variation of our method with 8 PoIs has the lowest errors in the ALL and PLL detection but significantly higher errors in transforming ITL (32.07 mm) and SSL (84.90 mm) ligament groups, suggesting that reducing the number of PoIs may lead to increased error in the ligaments that are not located on the vertebral body. The overall average error jumps to 19.49 mm with an RMSE of 6.14 mm, showing a decrease in the registration precision.
While the selected baseline outperforms our method in detecting the ligament points located in the posterior part of the vertebra (except ITL and SSL), it requires almost 25 times more computational effort (77.6 seconds per vertebra). Both variations of our proposed pipeline are significantly faster than ALPACA detecting landmarks at approx. 3.0 seconds per vertebra. 
\\ 
An additional advantage of our proposed method is its ability to accurately localize ligament landmarks on damaged vertebrae. As illustrated in Fig. \ref{fig:sample_res}, despite the presence of significant graded fractures on the endplates and anterior cortex buckling \cite{wang2017identifying}, our method successfully projects the ALL and PLL landmarks onto the vertebral body. The ligament landmarks detected on a healthy L1-vertebra by using our method are shown in Fig. \ref{fig:sample_res_healthy}. 


\section{Conclusion and Future Work}

In this paper, we proposed an automated pipeline for detection of 3D spine ligaments on the vertebral models. 
The conducted experiments demonstrate a high accuracy of our method in calculating the landmarks located on the anterior part of the vertebra including ALL and PLL groups as well as some of the posterior ligaments such as ITL. While the baseline method leads in precision for certain ligament locations, our pipeline significantly enhances the time efficiency by rapidly detecting landmarks in approximately 3.0 seconds per vertebra, showing a substantial improvement over existing methods that may require considerably more time to achieve similar level of precision.

Although the projection of landmarks onto the detected 3D model works effectively for most ligament groups such as the anterior longitudinal ligament (ALL), posterior longitudinal ligament (PLL) and  intermuscular transverse ligament (ITL), there is room for improvement in detecting posterior ligaments, particularly in areas with fewer edges, such as the smoother regions of the facet joints. To enhance the localization precision of the capsular ligament (CL),  ligamentum flavum (LF) and interspinous ligament (ISL) landmarks, we plan to develop alternative projection strategies. These strategies will involve dividing the posterior vertebral surface in five parts, one for each ligament group,  and applying heuristic-based methods to project the registered landmarks onto the vertebral surface.

\bibliographystyle{IEEEtran}
\bibliography{ref}

\end{document}